%% file: iclr2025_conference.tex
\documentclass{article} 
\usepackage{iclr2025_conference,times}
\usepackage{graphicx}

\input{math_commands.tex}

\usepackage{soul} 
\usepackage{booktabs}
\usepackage{hyperref}
\usepackage{url}
\usepackage{multirow}
\usepackage{graphicx}
\usepackage{caption}

\newcommand{\name}{\textsc{Ultra3D}}

\title{\name: Efficient and High-Fidelity 3D Generation with Part Attention}

\iclrfinalcopy 

\author{
        Yiwen Chen$^{1,2}$\thanks{Work done during a research internship at Math Magic.},~
        Zhihao Li$^{1}$,
        Yikai Wang$^{3}$,\\
        \textbf{
        Hu Zhang$^{2}$,
        Qin Li$^{2,4}$,
        Chi Zhang$^{5\dagger}$,
        Guosheng Lin$^{1}$\thanks{Corresponding Authors.}}
    \\
    {\normalsize $^{1}$Nanyang Technological University}~~~~
    {\normalsize $^{2}$Math Magic}~~~~
    {\normalsize $^{3}$Tsinghua University}~~~~\\
    {\normalsize $^{4}$School of Artificial Intelligence, Beijing Normal University}
    {\normalsize $^{5}$Westlake University}
    \\
    \tt \color{red}{\href{https://buaacyw.github.io/ultra3d/}{https://buaacyw.github.io/ultra3d/}}
}

\begin{document}

\maketitle

\input{0_abstract}

\input{1_intro}
\input{2_related}
\input{3_Pre}

\input{4_Method}

\input{5_exp}

\input{6_conclu}

\bibliography{iclr2025_conference}
\bibliographystyle{iclr2025_conference}

\end{document}

%% file: math_commands.tex
\usepackage{amsmath,amsfonts,bm}

\def\eqref#1{equation~\ref{#1}}

\def\1{\bm{1}}

\DeclareMathAlphabet{\mathsfit}{\encodingdefault}{\sfdefault}{m}{sl}
\SetMathAlphabet{\mathsfit}{bold}{\encodingdefault}{\sfdefault}{bx}{n}



%% file: 0_abstract.tex
\input{figs/teaser}
\begin{abstract}

Recent advances in sparse voxel representations have significantly improved the quality of 3D content generation, enabling high-resolution modeling with fine-grained geometry. However, existing frameworks suffer from severe computational inefficiencies due to the quadratic complexity of attention mechanisms in their two-stage diffusion pipelines. In this work, we propose \name, an efficient 3D generation framework that significantly accelerates sparse voxel modeling without compromising quality. Our method leverages the compact VecSet representation to efficiently generate a coarse object layout in the first stage, reducing token count and accelerating voxel coordinate prediction. To refine per-voxel latent features in the second stage, we introduce Part Attention, a geometry-aware localized attention mechanism that restricts attention computation within semantically consistent part regions. This design preserves structural continuity while avoiding unnecessary global attention, achieving up to a 6.7× speed-up in latent generation. To support this mechanism, we construct a scalable part annotation pipeline that converts raw meshes into part-labeled sparse voxels. 
Extensive experiments demonstrate that \name~supports high-resolution 3D generation at 1024 resolution and achieves state-of-the-art performance in both visual fidelity and user preference.

\end{abstract}

%% file: figs/teaser.tex
\vspace{-6mm}
\begin{figure}[h]
  \centering
  \includegraphics[width=\linewidth]{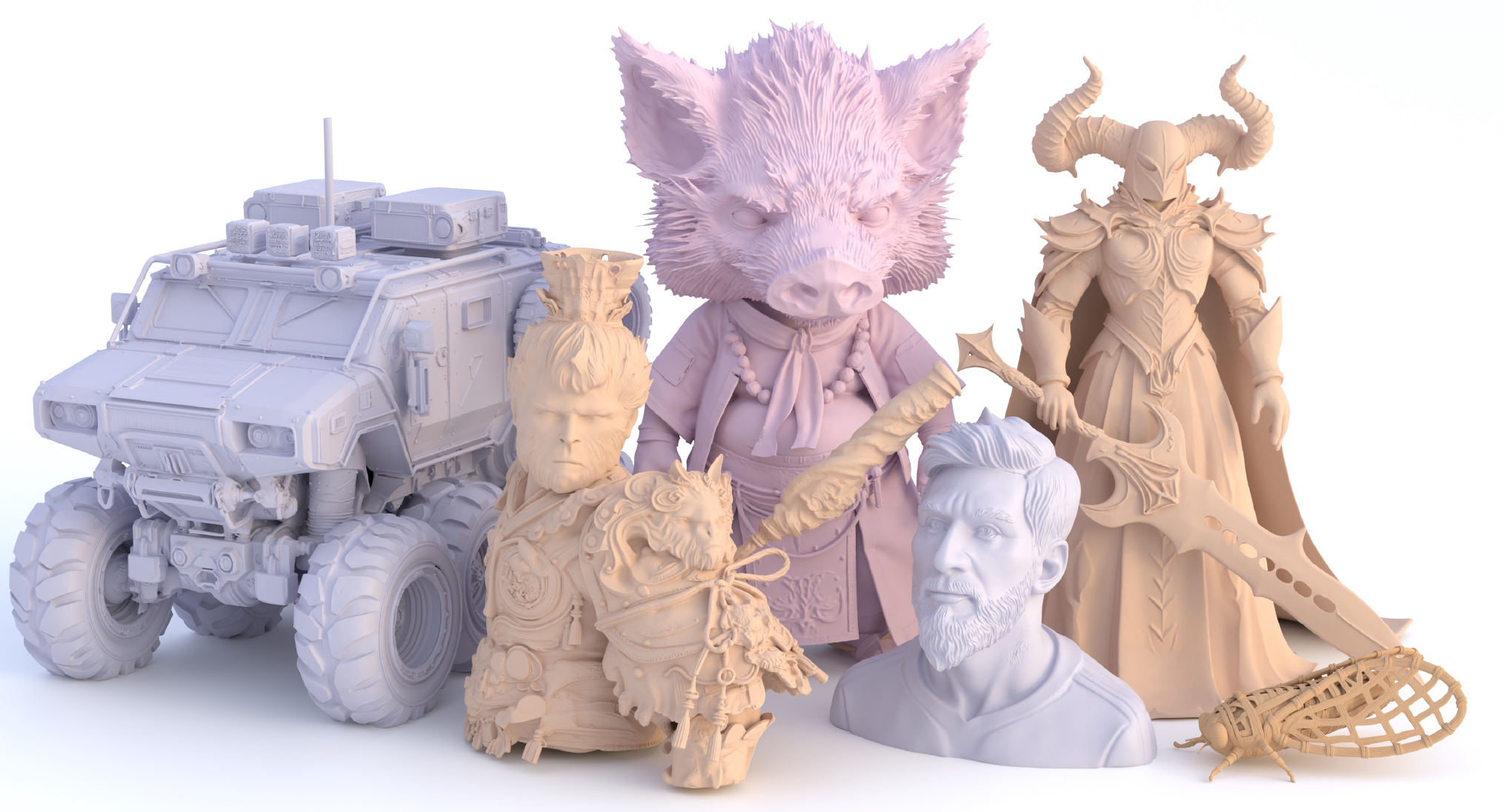}
  \caption{
    \textbf{Image-to-3D Generation Results of \name.} \name~delivers high-quality 3D meshes with fine-grained geometric details while maintaining efficient generation. Please zoom in to view detailed geometry.
  }
  \label{fig:teaser}
\end{figure}

%% file: 1_intro.tex
\section{Introduction}

Recent advances in generative modeling~\citep{dreamfusion,liu2023zero,hong2023lrm,zhang20233dshape2vecset,xiang2025structured3dlatentsscalable} have significantly expanded the frontier of 3D content creation.
These models enable the generation of high-resolution and structurally consistent 3D assets, supporting a wide array of applications in gaming, augmented and virtual reality (AR/VR), digital content creation, and robotics. The growing demand for scalable and controllable 3D generation pipelines in these domains has spurred extensive research into expressive representations and generative models~\citep{zhang20233dshape2vecset,xiang2025structured3dlatentsscalable} that can produce high-fidelity 3D content.

A notable advancement in this direction is the introduction of sparse voxel-based representations~\citep{ren2024xcube,xiang2025structured3dlatentsscalable}. Sparse voxel-based representations have emerged as powerful 3D representations due to their ability to capture fine-grained geometry~\citep{ren2024xcube,xiang2025structured3dlatentsscalable,ye2025hi3dgen,he2025triposf,wu2025direct3ds2gigascale3dgeneration,li2025sparc3dsparserepresentationconstruction}. In this paradigm, a 3D object is encoded as a sparse voxel grid, where each active voxel is associated with a latent feature vector. 
This design enables two complementary benefits: (1) the voxel grid provides a coarse yet globally consistent structural layout; and (2) the per-voxel latent features support localized surface modeling with fine granularity. Together, these attributes facilitate high-quality 3D generation with fine-grained geometry quality~\citep{xiang2025structured3dlatentsscalable}.
As a result, sparse voxels have been adopted by a series of state-of-the-art 3D generation frameworks~\citep{xiang2025structured3dlatentsscalable,ye2025hi3dgen,wu2025direct3ds2gigascale3dgeneration,he2025triposf,li2025sparc3dsparserepresentationconstruction}, becoming a mainstream solution for 3D modeling.

However, the benefits of this expressive representation come at the cost of computational efficiency. Most existing frameworks adopt a two-stage pipeline introduced by~\cite{xiang2025structured3dlatentsscalable}: first, predicting the coordinates of active voxels, followed by generating per-voxel latent vectors. Both stages are typically implemented using Diffusion Transformer (DiT)~\citep{peebles2023scalable}. As resolution increases, the number of tokens processed in each stage grows significantly, leading to substantial memory and computation overhead due to the quadratic complexity of attention computation~\citep{vaswani2017attention}. Consequently, current approaches are often constrained to low resolutions and limited output quality.

Built upon the success of sparse voxel generation frameworks~\citep{xiang2025structured3dlatentsscalable}, we aim to develop an efficient 3D generation pipeline that overcomes their efficiency limitations and supports high-resolution modeling. We observe that the two stages in the pipeline serve distinct purposes: the first stage constructs a coarse object layout, while the second stage refines fine surface details. In the first stage, prior methods typically compress sparse voxel coordinates into a dense feature grid at one-fourth the resolution for DiT modeling. Although this reduces the number of tokens, the computation remains costly at high resolutions. This is because directly predicting tens of thousands of coordinates is inherently complex and computationally intensive. 

Therefore, we depart from direct prediction and instead propose using more efficient methods to first generate a coarse mesh, which is then voxelized to produce the sparse voxel coordinates. Notably, since the coarse mesh is only expected to convey the overall structure rather than fine surface details, it can be efficiently represented using a compact 3D representation. Specifically, we adopt VecSet~\citep{zhang20233dshape2vecset}, an efficient 3D representation that encodes each 3D asset into a small set of latent tokens, for generating sparse voxel coordinates. While VecSet is less expressive for fine-grained geometry than sparse voxels, its compactness enables highly efficient generation, which is sufficient for producing the coarse object layout without compromising final generation quality. In practice, this reduces generation time from several minutes to just a few seconds for producing 128-resolution sparse voxels.

\input{figs/window_attn}

To refine the coarse outputs from the first stage with rich and precise latent vectors, previous methods typically adopt a DiT architecture with expensive full attention~\citep{xiang2025structured3dlatentsscalable,ye2025hi3dgen,li2025sparc3dsparserepresentationconstruction}. However, we observe that the sparse voxels predicted from the first stage already captures the overall object structure, making global attention often redundant and inefficient for local detail refinement. Motivated by this, we introduce \textbf{Part Attention}, a geometry-aligned, localized attention mechanism tailored for sparse voxels. It leverages voxel-level part annotations to restrict attention computation within each part group, thereby significantly enhancing efficiency by avoiding unnecessary interactions across unrelated regions.

Compared to the widely used window attention in LLMs~\citep{beltagy2020longformer,zaheer2021bigbirdtransformerslonger}, Part Attention is better suited for 3D sparse voxels. As shown in Fig.~\ref{fig:window_attn}, fixed window attention struggles with the irregular nature of sparse voxels, often leading to inconsistent styles and degraded performance. In contrast, Part Attention partitions tokens according to the object’s parts and geometric layout, respecting semantic boundaries and object structure, thereby preserving consistency and high-quality output. Our experiments demonstrate that Part Attention achieves up to a 6.7× speed-up without compromising generation quality.

To support Part Attention, we construct an efficient, large-scale part annotation pipeline. Given the limited availability of part-labeled 3D datasets~\citep{yang2024sampart3d}, we adopt PartField~\citep{liu2025partfield} as our base segmentation model to convert raw mesh data into sparse voxels with part annotations. To ensure data quality, we further apply a series of lightweight filtering strategies. Our pipeline takes only a few seconds to process each mesh with high annotation quality, making it practical for large-scale dataset annotation.

Our experiments show that our design accelerates the generation pipeline by 3.3× without compromising quality. Furthermore, \name~achieves state-of-the-art performance in both visual fidelity and user preference.

Our main contributions are summarized as follows:
\begin{itemize}
    \item We present \textbf{\name}, an efficient high-resolution 3D generation framework that first produces sparse voxels via the compact VecSet representation~\citep{zhang20233dshape2vecset} and then refines them through per-voxel latent generation~\citep{xiang2025structured3dlatentsscalable}, enabling both speed and fidelity.
    \item We introduce \textbf{Part Attention}, an efficient localized attention mechanism tailored for sparse voxels that performs attention computation independently within each part group. By preserving geometric continuity, it achieves up to 6.7$\times$ speed-up without compromising quality.
    \item We develop a scalable part annotation pipeline that efficiently converts raw meshes into part-labeled sparse voxels, enabling high-quality annotations at scale.
  \item Our experiments demonstrate that \name~achieves state-of-the-art performance in both visual quality and user preference, while achieving up to 3.3× speed-up over baseline methods without compromising generation quality.
\end{itemize}

%% file: figs/window_attn.tex
\begin{figure}[h]
  \centering
   \includegraphics[width=\linewidth]{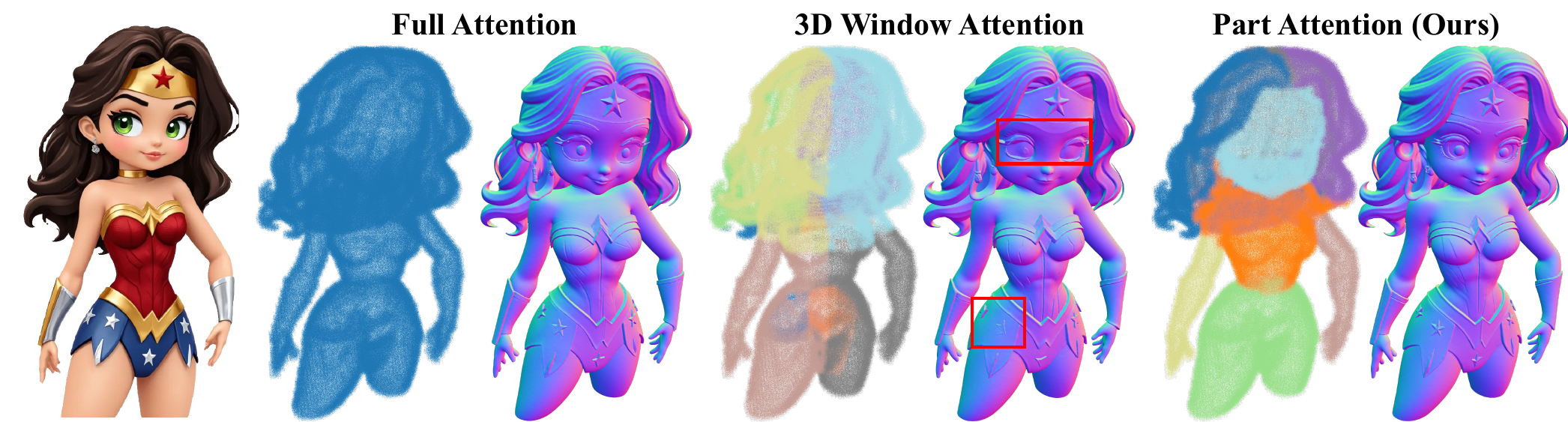}
   \caption{
\textbf{Expeiments on different attention mechanisms.} Each color denotes an attention group, within which attention is computed independently. All other settings remain unchanged, with only the attention mechanism being replaced. 3D Window Attention partitions the object space into 8 fixed regions by splitting at the center along each axis. This fixed partitioning often misaligns with semantic boundaries, leading to degraded quality and style inconsistencies. }
\vspace{-5mm}
   \label{fig:window_attn}
\end{figure}

%% file: 2_related.tex
\section{Related Work}

The field of 3D object generation has witnessed rapid progress in recent years~\citep{liu2023zero, dreamfusion,DBLP:conf/cvpr/WangDLYS23, liu2023one2345,prolificdreamer, DBLP:conf/iccv/ChanNCBPLAMKW23, liang2024luciddreamer,DBLP:conf/cvpr/Lin0TTZHKF0L23, yi2024gaussiandreamer, chen2023text,long2023wonder3d, DBLP:conf/iclr/ShiWYMLY24, DBLP:conf/iclr/LiuLZLLKW24,DBLP:journals/corr/abs-2312-02201, DBLP:journals/corr/abs-2405-20343, DBLP:journals/corr/abs-2405-11616, tang2023dreamgaussian, hong2023lrm,tang2024lgm,xu2023dmv3d,wang2023pf,li2023instant3d,zhang2024gs,zou2024triplane,wei2024meshlrm,xu2024instantmesh,DBLP:journals/corr/abs-2403-02234,tochilkin2024triposr}, with several promising directions emerging for scalable 3D object generation~\citep{zhang20233dshape2vecset,zhang2024clay,xiang2025structured3dlatentsscalable,xiong2025octfusionoctreebaseddiffusionmodels,nash2020polygen,siddiqui2024meshgpt,chen2024meshanythingartistcreatedmeshgeneration}. We broadly categorize current methods into three main categories.

\subsection{Vector Set-based Object Generation} 3DShape2Vecset~\citep{zhang20233dshape2vecset} introduces a pipeline that uses a VAE to compress 3D shapes into a compact latent space named Vector Set (VecSet), and subsequently trains a diffusion model in this latent space. Follow-up works~\citep{zhang2024clay,zhao2024michelangelo,wu2024direct3d,li2024craftsman,lan2024ln3diff,DBLP:journals/corr/abs-2403-02234, li2025triposg, chen2025meshgengeneratingpbrtextured, zhao2025hunyuan3d20scalingdiffusion,hunyuan3d2025hunyuan3d21imageshighfidelity, lai2025hunyuan3d25highfidelity3d,yang2025holopartgenerative3damodal,lin2025partcrafterstructured3dmesh,tang2025efficientpartlevel3dobject} have demonstrated that this pipeline is highly scalable and capable of producing high-resolution meshes from large-scale datasets. Due to the compact nature of the VecSet representation (typically only a few thousand tokens), both training and inference are computationally efficient. \cite{yang2025holopartgenerative3damodal,lin2025partcrafterstructured3dmesh, tang2025efficientpartlevel3dobject} have shown the flexibility of this representation in integrating part information, enabling simultaneous generation of 3D object and part-label. These methods can be combined with our work, as they can provide high-quality part labels for our Part Attention mechanism. While VecSet is flexible and efficient, it falls short in modeling fine-grained 3D surface details when compared to sparse voxel-based methods~\citep{xiang2025structured3dlatentsscalable,he2025triposf,li2025sparc3dsparserepresentationconstruction}.

\subsection{Sparse Voxel-based Object Generation} Trellis~\citep{xiang2025structured3dlatentsscalable} introduces a novel 3D representation known as the structured latent, which encodes 3D assets as 3D sparse voxels augmented with latent vectors to capture fine-grained surface details. The generation of this representation typically follows a two-stage pipeline. In the first stage, the voxel coordinates are compressed into a low-resolution continuous feature grid using a lightweight 3D convolutional VAE, and then a Diffusion Transformer (DiT)~\citep{peebles2023scalable} is trained to generate this grid from noise. In the second stage, another DiT generates the corresponding latent features, conditioned on these voxel coordinates. Both stages are trained independently using the conditional flow matching objective~\citep{lipman2023flow}. Subsequent studies~\citep{ye2025hi3dgen,he2025sparseflexhighresolutionarbitrarytopology3d, wu2025direct3ds2gigascale3dgeneration, li2025sparc3dsparserepresentationconstruction} demonstrate that sparse voxel excels at modeling extremely fine-grained 3D geometry, significantly outperforming prior representations in terms of geometric fidelity. While sparse voxel provides superior fidelity, it suffers from significant computational overhead. As the resolution increases, the number of sparse voxel tokens can exceed 20K, leading to expensive attention computation.

\subsection{Autoregressive Mesh Generation}
MeshGPT~\citep{siddiqui2024meshgpt} introduced a vertex-by-vertex autoregressive generation approach for meshes, producing outputs that closely resemble those created by human artists—making them particularly valuable in applications such as gaming and digital content creation. Building on this direction, MeshAnything~\citep{chen2024meshanythingartistcreatedmeshgeneration} proposed a scalable shape-to-mesh training setting focused on topology generation. Subsequent developments~\citep{chen2024meshxl,tang2024edgerunnerautoregressiveautoencoderartistic,chen2024meshanythingv2artistcreatedmesh,weng2024scalingmeshgenerationcompressive,hao2024meshtronhighfidelityartistlike3d,wang2024llamameshunifying3dmesh,zhao2025deepmeshautoregressiveartistmeshcreation}, demonstrated that with large-scale training data, such models can generate highly detailed meshes with thousands of faces. This line of work stands out for its ability to mimic artist-created meshes, though it also faces challenges related to high computational cost due to the large number of autoregressive tokens.

%% file: 3_Pre.tex
\section{Preliminaries}
\label{sec:preliminary}

\subsection{Sparse Voxel-based Representations}
\label{sec:preliminary:sparse_voxel}

\input{figs/pip}

Introduced by~\cite{xiang2025structured3dlatentsscalable}, sparse voxel-based representations~\citep{xiang2025structured3dlatentsscalable,ye2025hi3dgen,he2025triposf,li2025sparc3dsparserepresentationconstruction} compress a 3D object into a set of latent features distributed over a sparse voxel grid. Formally, a shape is encoded as a collection of tuples $\{(\boldsymbol{z}_i,\boldsymbol{p}_i)\}_{i=1}^{L}$, where $\boldsymbol{p}_i \in \{0, \ldots, N{-}1\}^3$ denotes the coordinate of an active voxel in a cubic grid of resolution $N$, and $\boldsymbol{z}_i \in \mathbb{R}^C$ is the latent vector associated with that voxel. This formulation preserves both geometric fidelity and spatial locality, making it well-suited for high-resolution surface modeling.

The generation of this representation typically follows a two-stage pipeline~\citep{xiang2025structured3dlatentsscalable}. In the first stage, the coordinates $\{\boldsymbol{p}_i\}_{i=1}^{L}$ are compressed into a low-resolution continuous feature grid $\boldsymbol{S} \in \mathbb{R}^{D \times D \times D \times C}$ using a lightweight 3D convolutional VAE. A DiT~\citep{peebles2023scalable} is trained to generate $\boldsymbol{S}$ from noise. In the second stage, another DiT conditioned on $\{\boldsymbol{p}_i\}_{i=1}^{L}$ generates the corresponding latent features $\{\boldsymbol{z}_i\}_{i=1}^{L}$. Both stages are trained independently using the conditional flow matching (CFM) objective~\citep{lipman2023flow}.

\subsection{Vector Set Representation}
\label{sec:preliminary:vecset}

The Vector Set (VecSet)~\citep{zhang20233dshape2vecset} representation is a compact and permutation-invariant representation for 3D object modeling, designed to facilitate scalable generative pipelines. VecSet encodes a 3D object as an unordered set of latent vectors, where each latent vector encodes localized 3D shape and semantic attributes. With only a few thousand vectors per shape, VecSet supports fast diffusion-based generation, scaling well to high-resolution 3D synthesis~\citep{zhang20233dshape2vecset,zhang2024clay,zhao2025hunyuan3d20scalingdiffusion}.

While sparse voxels offer strong expressiveness and state-of-the-art surface fidelity~\citep{xiang2025structured3dlatentsscalable,he2025triposf,li2025sparc3dsparserepresentationconstruction}, their large token count at high resolutions poses significant challenges for efficient training and inference. In contrast, the compactness of VecSet~\citep{zhang20233dshape2vecset} enables fast generation with only a few thousand latent tokens per shape, but its lack of explicit spatial structure limits its ability to capture fine-grained geometry, making it less suitable for high-resolution surface modeling.

%% file: figs/pip.tex
\begin{figure}[h]
  \centering
   \includegraphics[width=\linewidth]{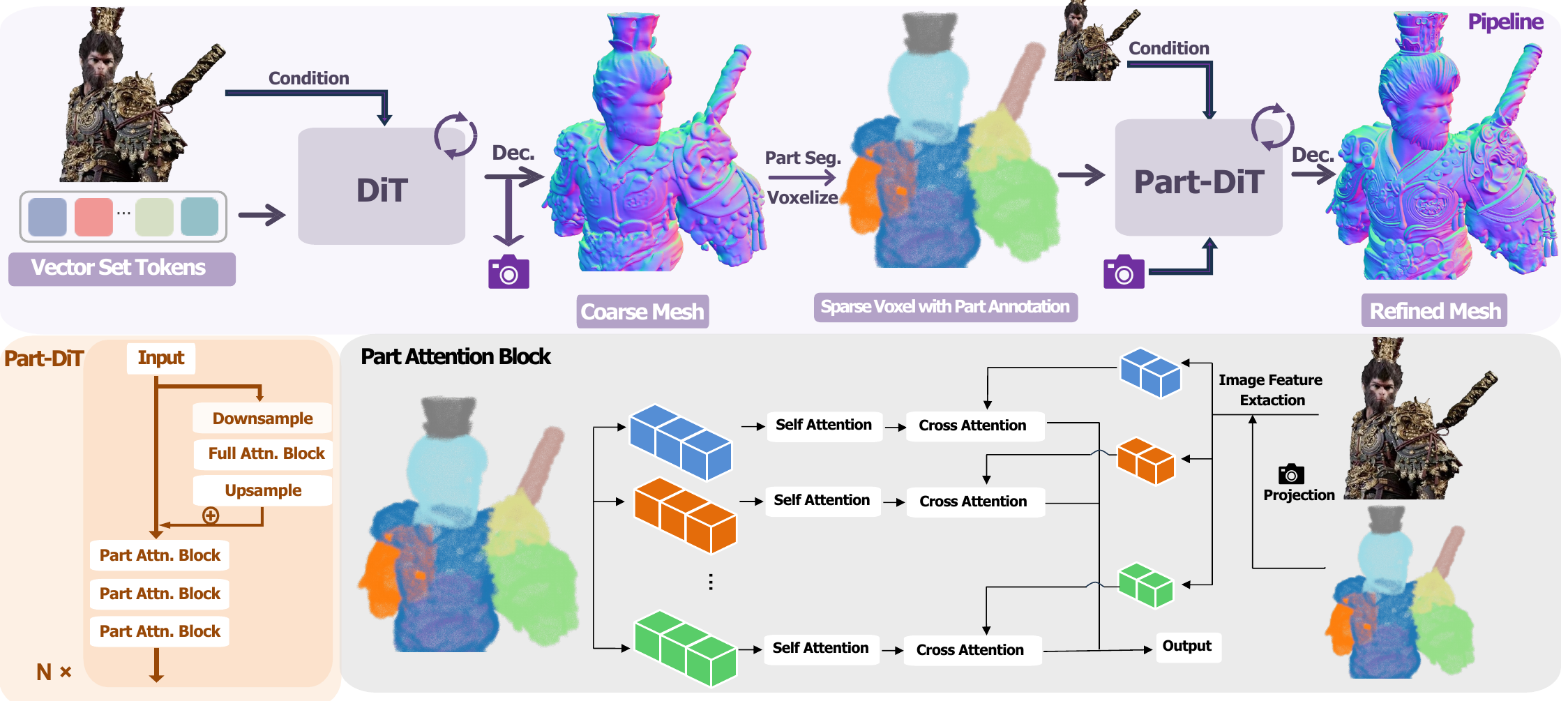}
\caption{
    \textbf{Pipeline Overview.} We introduce \name, an efficient and high-quality 3D generation framework that first generates sparse voxel layout via VecSet and then refines it by generating per-voxel latent. The core of \name~is Part Attention, an efficient localized attention mechanism that performs attention computation independently within each part group. Besides, when the input condition is an image, each part group performs cross attention only with the image tokens onto which its voxel tokens are projected.}
\vspace{-4mm}
   \label{fig:pip}
\end{figure}

%% file: 4_Method.tex
\section{Method}

We present \name, an efficient 3D generation framework that supports high-resolution and high-fidelity modeling. It first produces sparse voxels using the compact VecSet representation~\citep{zhang20233dshape2vecset}, and then refines them through per-voxel latent generation~\citep{xiang2025structured3dlatentsscalable}, achieving both speed and fidelity.

At the core of \name~is \textbf{Part Attention}, an efficient attention mechanism designed to accelerate sprase voxel generation by performing attention computation independently within each part group (see Sec.~\ref{method:part_attention}). We describe the overall generation pipeline in Sec.~\ref{sec:overview}, and introduce our scalable part annotation pipeline in Sec.~\ref{sec:preparation}.

\subsection{Part Attention}
\label{method:part_attention}

In the widely used sparse voxel generation pipeline~\citep{xiang2025structured3dlatentsscalable}, the primary computational bottleneck lies in generating the local latents $\{\boldsymbol{z}_i\}_{i=1}^{L}$. For instance, generating a 1024-resolution mesh typically requires performing attention over sparse voxel at resolution 64 or 128. In these cases, the average number of active voxels reaches approximately 20K and 60K, respectively. This makes attention computation prohibitively expensive due to its quadratic scaling with respect to token count. Moreover, as the mesh resolution increases, the corresponding sparse voxel becomes more detailed and effectively captures the global shape and topology of the object. Consequently, the role of sparse voxel latent generation shifts toward refining local surface geometry rather than modeling the overall structure. These observations suggest that full global attention across all tokens is both unnecessary and inefficient.

A straightforward solution is to introduce more efficient attention mechanisms, such as window attention, as explored in the context of large language models~\citep{beltagy2020longformer}. However, as illustrated in Fig.~\ref{fig:window_attn}, these methods cannot be directly applied to our case. These attention strategies often adopt a certain fixed pattern and leads to attention boundaries that misalign with the semantic layout of the object, leading to inconsistent styles and degraded performance. We find that the fundamental reason is the structural complexity of 3D assets: unlike text sequences, 3D sparse voxel cannot be easily divided into semantically meaningful local blocks using fixed partitioning schemes. 

The above observations motivate the introduction of a new 3D attention design, \textbf{Part Attention}. Part Attention performs grouping based on externally provided part information. Specifically, for each active voxel $\boldsymbol{p}_i$, we assign a part index $a_i \in \{1, \ldots, K\}$, where $K$ denotes the number of part groups. Based on this grouping, we apply attention computations that are restricted within each part. Part Attention respects semantic boundaries and object structure, thereby preserving consistency and high-quality output. We next illustrate the designs of Part Self Attention and Part Cross Attention.

\textbf{Part Self Attention.} During self-attention computation, each token only attends to other tokens that belong to the same part. Let $\text{Attn}(i, j)$ denote the attention mask value from the token located at $\boldsymbol{p}_i$ to the token at $\boldsymbol{p}_j$. The attention is masked as:

\begin{equation}
\text{Attn}(i, j) = 0 \quad \text{if} \quad a_i \ne a_j.
\end{equation}

This masking strategy enforces part-level locality by restricting attention to within each part group, enabling more structured and efficient attention computation. Under the assumption that each part group contains a similar number of tokens, the computation cost of attention is reduced by nearly a factor of $K$.

\input{figs/reso_compare}

\textbf{Part Cross Attention.} 
In the image-to-3D setting, cross-attention between the 3D sparse voxel and the image features becomes increasingly expensive at high resolutions. To reduce this cost, we also leverage the part index $a_i \in \{1, \ldots, K\}$ to constrain attention across modalities.

Our goal is to allow each 3D voxel to only interact with 2D image regions that correspond to the same part. To implement this, we project each 3D part group onto the condition image using externally provided camera parameters, and assign the part index to the pixels onto which the 3D voxels are projected. Since multiple 3D parts may project onto the same pixel, each 2D image token is associated with a set of part indices.

Let $\text{Attn}(i, j)$ denote the attention mask value from the voxel token located at $\boldsymbol{p}_i$ to the image token indexed by $j$, and let $\mathcal{A}_j$ denote the set of part indices assigned to the $j$-th image token. The attention is masked as:

\begin{equation}
\text{Attn}(i, j) = 0 \quad \text{if} \quad a_i \notin \mathcal{A}_j,
\end{equation}

ensuring that each voxel token only attends to pixels aligned with its own part group. This masking reduces the overall cross-attention cost while preserving semantic consistency between 3D and 2D tokens.

In \name, the part labels required by Part Attention are provided by PartField~\citep{liu2025partfield}during both training and inference. For camera information, we use ground-truth cameras during training and obtain estimated cameras from the VecSet decoder during inference.

\subsection{\name}
\label{sec:overview}
Following \cite{xiang2025structured3dlatentsscalable}, our generation pipeline is divided into two stages: the first stage generates sparse voxel coordinates that capture the overall object structure, and the second stage refines these coordinates by producing high-quality per-voxel latent features for final mesh reconstruction.

\textbf{VecSet-based Sparse Voxel Generation.} In our initial experiments, we adopted the sparse voxel generation architecture from Trellis~\citep{xiang2025structured3dlatentsscalable}, as outlined in Sec.~\ref{sec:preliminary:sparse_voxel}. However, we found that this design struggles to balance quality and efficiency at high resolutions. As shown in Fig.~\ref{fig:reso_compare}, when aiming to generate meshes at a resolution of 1024, high-quality results typically require sparse voxels at 128 resolution. Under Trellis~\citep{xiang2025structured3dlatentsscalable} formulation, this corresponds to applying diffusion-based generation over a $32^3$ dense grid—equivalent to full attention over 32K tokens—which incurs prohibitive computational cost during both training and inference. Alternatively, downsampling to a $16^3$ grid reduces the token count but leads to a significant drop in generation quality, making it inadequate for high-fidelity outputs.

Faced with this dilemma, we turn to VecSet~\citep{zhang20233dshape2vecset} for its compact representation and generation efficiency, making it a suitable choice for modeling coarse object structures at scale. As the vector set typically contains only a few thousand tokens, its computational cost is significantly lower than that of sparse voxel-based methods. Specifically, we first use it to generate a 512-resolution mesh, which is then voxelized into sparse voxel. Since the sparse voxel typically has a resolution of 64 or 128, the surface fidelity limitations of VecSet-based models are largely mitigated and have minimal impact on downstream quality.

\input{figs/compare_label}

As illustrated in Sec.~\ref{sec:preliminary:vecset}, our implementation of VecSet~\citep{zhang20233dshape2vecset} generation pipeline follows~\cite{zhao2025hunyuan3d20scalingdiffusion}. The shape VAE encodes point clouds sampled from mesh surfaces into a fixed-length vector set, and decodes them into Sign Distance Function (SDF). In addition, since Part Cross Attention requires camera information, we also encode the camera parameters into the latent vector.

\textbf{Sparse Latent Generation.} Despite the efficiency of VecSet~\citep{zhang20233dshape2vecset}, its inability to capture fine surface geometry necessitates a subsequent refinement stage using sparse voxel-based method. We follow Trellis~\citep{xiang2025structured3dlatentsscalable} and adopt a 3D sparse VAE, along with its corresponding flow-matching DiT, to generate per-voxel latents $\{\boldsymbol{z}_i\}_{i=1}^{L}$. We adopt the Sparconv-VAE introduced by \cite{li2025sparc3dsparserepresentationconstruction} as our 3D sparse VAE. It compresses 3D shapes into sparse voxel with latents $\{(\boldsymbol{z}_i, \boldsymbol{p}_i)\}_{i=1}^{L}$ as illustrated in Sec.~\ref{sec:preliminary:sparse_voxel}.

For DiT modeling, we replace most attention blocks with Part Attention to improve efficiency. A small number of full attention layers are retained to align styles across part groups. However, since full attention at high resolutions is computationally expensive, we introduce a residual block that performs full attention at a lower resolution. Specifically, as shown in Fig.~\ref{fig:pip}, we downsample the sparse voxels, apply full attention, and then upsample the features to fuse them back—enabling efficient cross-part communication at low cost. In practice, we stack one full-attention block followed by three Part Attention blocks repeatedly to build the DiT architecture.

The part labels are provided by an external part annotation model~\citep{liu2025partfield} during both training and inference. Note that recent advances in VecSet-based part-aware 3D generation~\citep{yang2025holopartgenerative3damodal,lin2025partcrafterstructured3dmesh, tang2025efficientpartlevel3dobject} offer an alternative for directly generating sparse voxels with part annotations, which can be integrated into our framework.

It is worth noting that, for annotation efficiency, we use a fixed 8-part grouping during training. However, this does not limit the model's flexibility at inference time, as it can accept inputs with varying numbers of part groups. As shown in Fig.~\ref{fig:compare_label}, our model consistently produces high-quality outputs even when the number of part groups differs from that used during training.

\input{figs/data}

\subsection{Sparse Voxel Part Annotation}
\label{sec:preparation}

In this section, we illustrate our data preprocessing pipeline for producing part annotations required by Part Attention. While existing large-scale datasets~\citep{chang2015shapenet, deitke2023objaverse, deitke2023objaversexl} provide a wealth of 3D assets, most of them lack part-level annotations. Therefore, we construct a part annotation pipeline that efficiently transforms raw meshes into sparse voxel with part annotations. 

Since our goal is to annotate millions of 3D assets, methods that require several minutes per shape are impractical. We therefore adopt PartField~\citep{liu2025partfield}, an efficient and high-quality part segmentation model. PartField is a feedforward network that takes a point-sampled 3D shape as input and predicts a feature field in the form of a triplane, which is subsequently clustered to obtain part-level segmentation. Its fast inference speed and reasonably accurate predictions make it well-suited for large-scale preprocessing.

To process raw mesh into sparse voxels with part annotations, we first uniformly sample point clouds from the mesh surface and input them into PartField~\citep{ liu2025partfield} to obtain a triplane feature field. We then query per-point features of the sampled point clouds from the triplane and voxelize the raw mesh. For each voxel, we average the features of the sampled points within it to produce a sparse voxel with part-aware voxel features. Finally, we apply Agglomerative Clustering to segment the voxels into parts based on these features~\citep{liu2025partfield}.

A key challenge during clustering is determining the appropriate number of part groups. Since Part Attention benefits from part segmentation that aligns with geometric structures, the ideal number of clusters can vary across different objects. Using too many clusters results in over-segmentation with noisy and fragmented parts, while too few clusters limit the computational advantage of Part Attention. Although one could design heuristics to adaptively determine the optimal number of clusters per object, we found this significantly increases the preprocessing cost. As a practical compromise, we apply a fixed number of clusters to all meshes and empirically set the number to 8. This setting yields reasonable part segmentations for the vast majority of samples.

After clustering, we apply two filtering criteria to discard samples with suboptimal segmentations. First, we heuristically assume that samples dominated by a single part group tend to exhibit poor segmentation quality. To identify such cases, we compute the voxel ratio of each part group, square these ratios, and sum them as a measure of imbalance. A higher score indicates a more uneven part distribution. Samples exceeding a fixed threshold are removed to eliminate those with overly dominant part groups. Second, well-segmented shapes typically exhibit strong spatial consistency, where most voxels are surrounded by neighbors that share the same part label. To quantify this, we define a metric called neighborhood inconsistency—the proportion of voxels whose
neighbors have different part labels. Samples with low full-neighborhood consistency are discarded, as they often reflect fragmented or noisy segmentations. With filtering included, the full preprocessing pipeline is highly efficient, processing each mesh in just 2 seconds on an A800 GPU. As shown in Fig.~\ref{fig:data}, the majority of processed samples exhibit high-quality segmentations, indicating the reliability of our annotation pipeline.

\input{figs/compare}

During inference, we first pass the input image through our sparse voxel generation pipeline to obtain a coarse mesh. We then apply the same part annotation procedure to obtain a part-labeled sparse voxel, which is required by the Part Attention.

%% file: figs/reso_compare.tex
\begin{figure}[h]
  \centering
   \includegraphics[width=\linewidth]{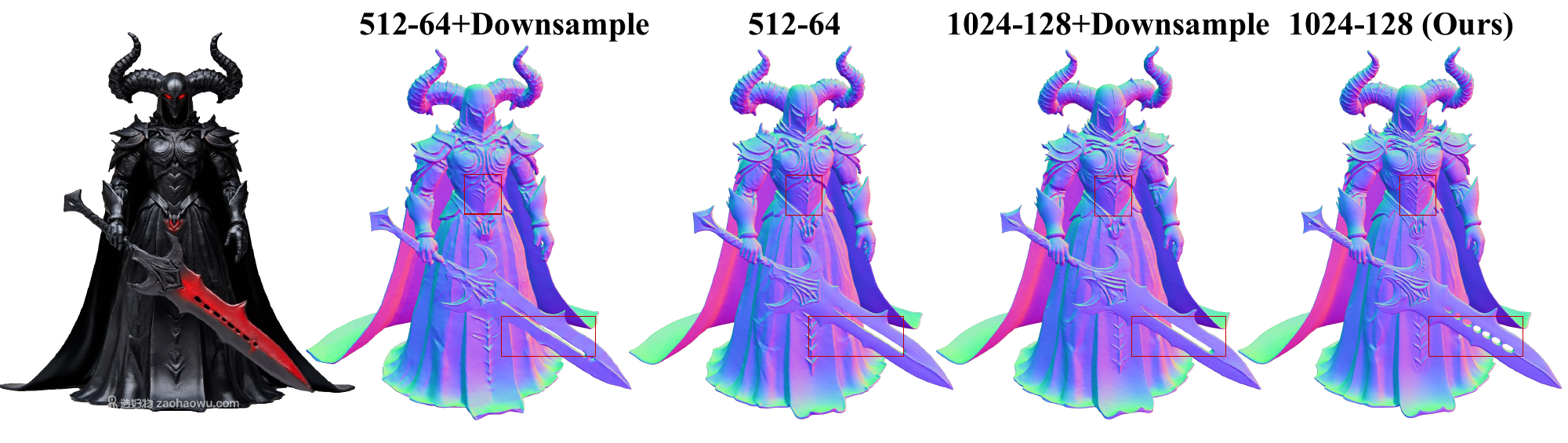}
\caption{
\textbf{Impact of Resolution on Generation Quality.} We compare results under different configurations, where “512\_64” denotes a mesh resolution of 512 and a sparse voxel resolution of 64. In previous works, to reduce computational cost in the second stage, the sparse voxels are typically downsampled by half before attention computation in the DiT, then upsampled afterward—annotated as “Downsample” in the figure. As shown, both the mesh resolution and the sparse voxel resolution used during attention computation significantly impact the final quality. However, due to efficiency constraints, prior methods were limited to lower resolutions. In contrast, our efficient framework supports higher sparse voxel resolutions, making high-quality generation feasible.
}
\label{fig:reso_compare}
\end{figure}

%% file: figs/compare_label.tex
\begin{figure}[h]
  \centering
   \includegraphics[width=\linewidth]{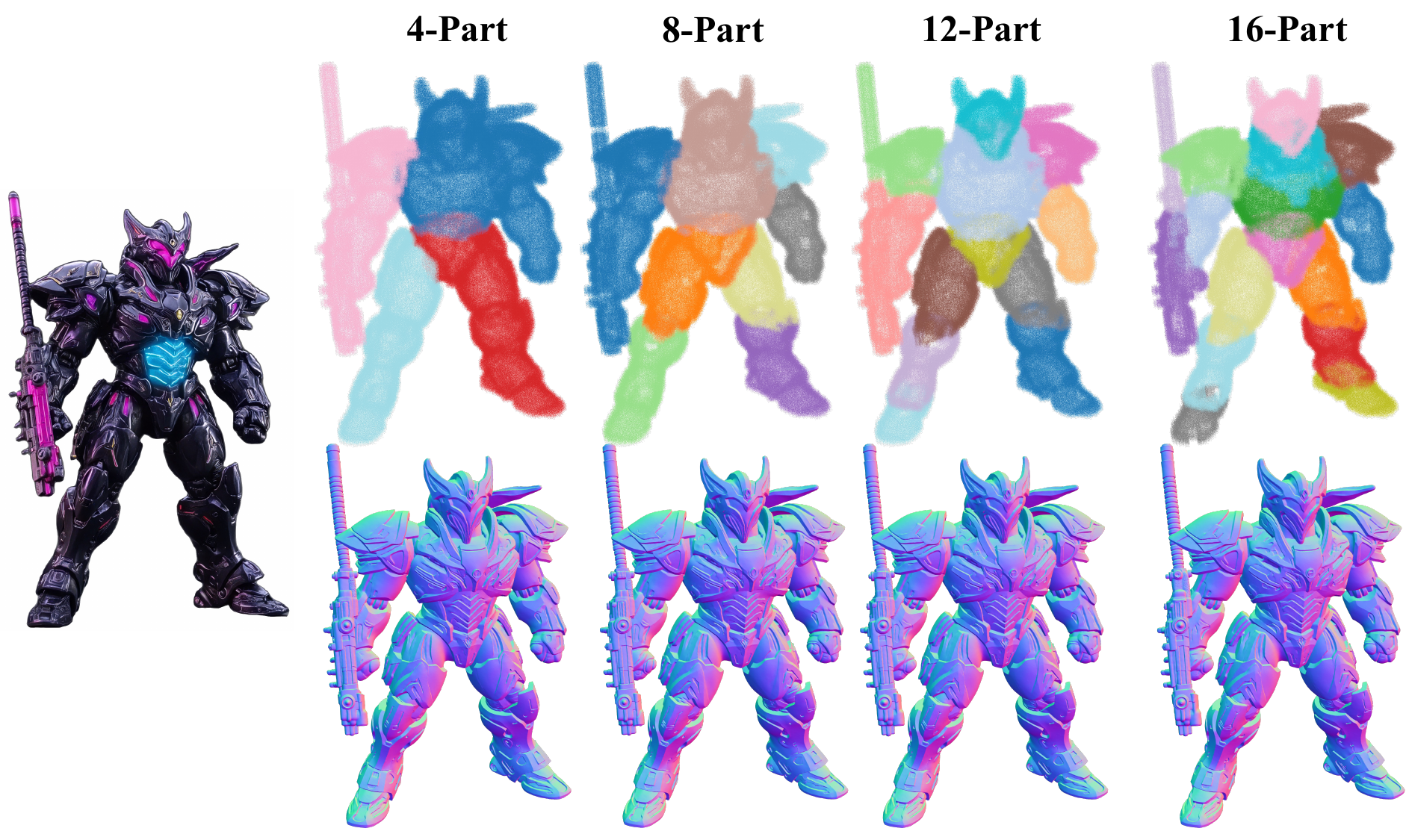}
\caption{
    \textbf{Robustness of Part Annotation}. Although our method is trained using data with exactly 8 part groups, we find it to be robust to variations in part annotation. Varying the number of part groups has little impact on generation quality, suggesting that increasing the number of annotated part groups can further accelerate computation without compromising performance.
}
   \label{fig:compare_label}
\end{figure}

%% file: figs/data.tex
\begin{figure}[h]
  \centering
   \includegraphics[width=\linewidth]{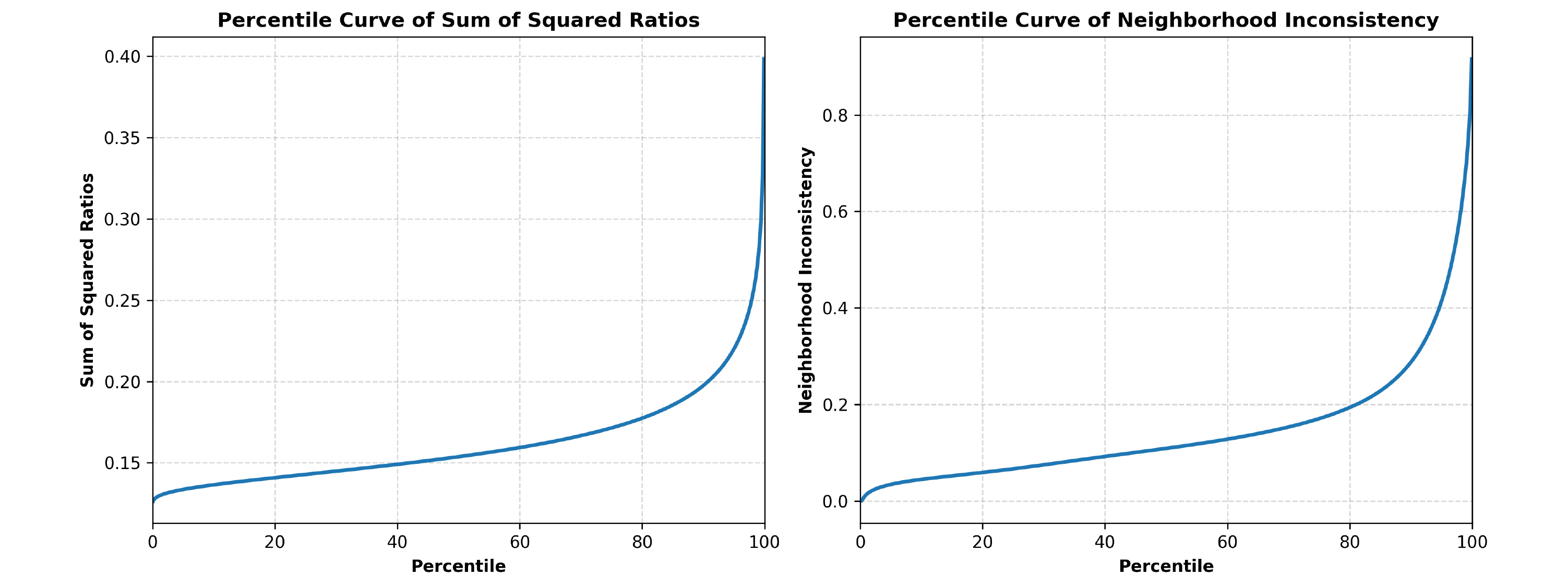}

\caption{
\textbf{Percentile for Filtering Metrics of Part Annotation.} 
We apply two metrics to filter poorly segmented samples: (1) the sum of squared voxel ratios, which identifies imbalanced part distributions, and (2) neighborhood inconsistency, which measures the proportion of voxels whose neighbors have different part labels. Lower values on both metrics indicate better segmentation quality. As shown in the plot, most samples exhibit stable and low values across both metrics, indicating high annotation quality.}
   \label{fig:data}
\end{figure}

%% file: figs/compare.tex
\begin{figure}[h]
  \centering
   \includegraphics[width=\linewidth]{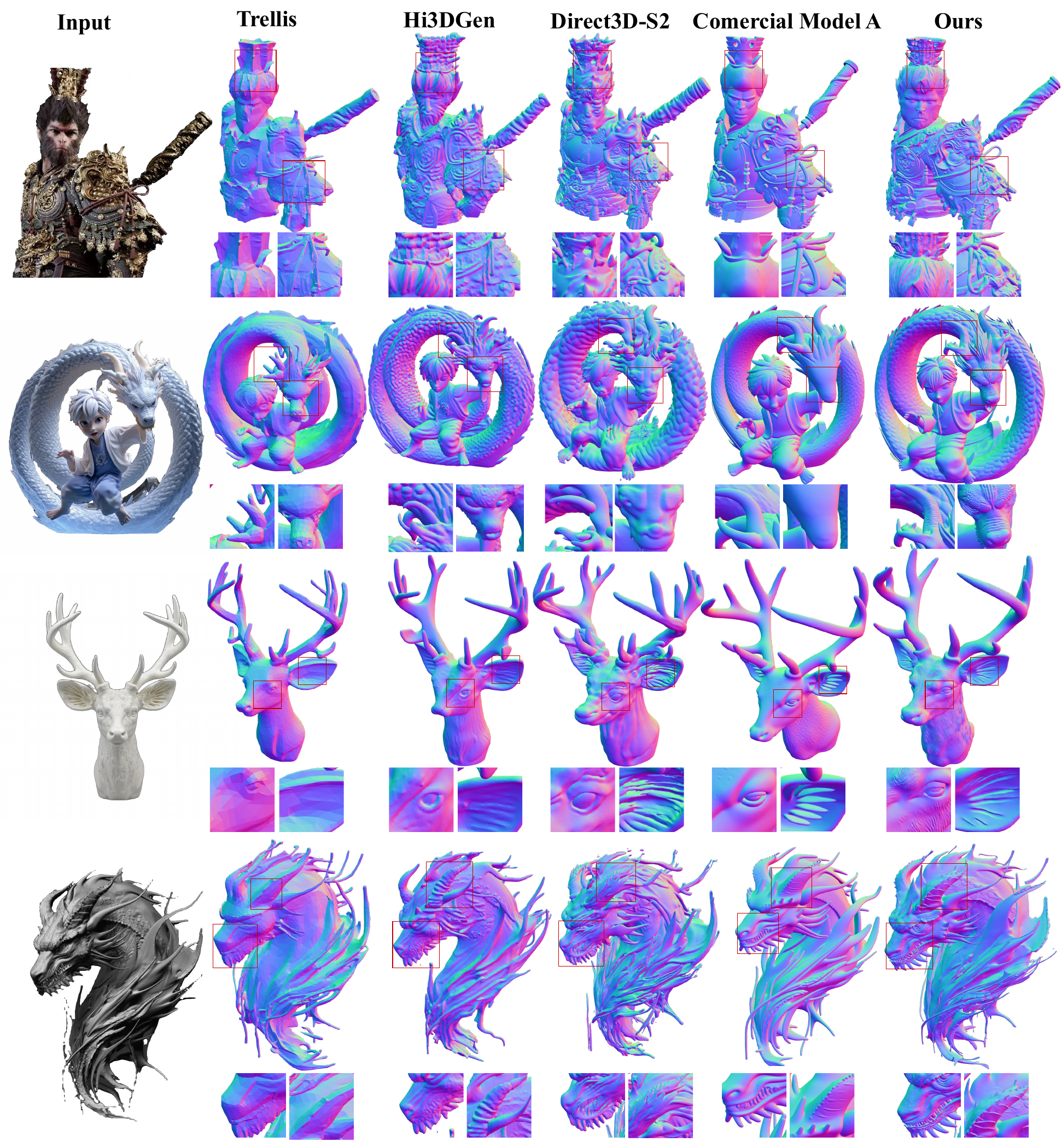}
\caption{
\textbf{Comparison with Prior Methods.} Our method produces higher fidelity and richer surface details. As highlighted in the red boxes, our results align more closely with the input image compared to other methods.
}
\vspace{-6mm}
   \label{fig:compare}
\end{figure}

%% file: 5_exp.tex
\section{Experiments}
\subsection{Implementation Details}
\name~adopts a sparse voxel resolution of 128 and produces output meshes at a resolution of 1024. Unless otherwise specified, all results reported in this paper are generated with this setting.
Our 3D sparse VAE adopts the settings from~\cite{li2025sparc3dsparserepresentationconstruction}. The VecSet-based generator follows the configuration of~\cite{zhao2025hunyuan3d20scalingdiffusion}, while our part-aware DiT is built upon~\cite{xiang2025structured3dlatentsscalable}, incorporating the modifications described in Sec.~\ref{sec:overview}, and scaling up the model size to 1.8B parameters.

All models are trained on a private 3D dataset, annotated using the part annotation pipeline detailed in Sec.~\ref{sec:preparation}. Following the procedure described in the method, we exclude samples where the sum of squared part ratios exceeds 25\% or the neighborhood inconsistency exceeds 25\%. The SparConv-VAE~\citep{li2025sparc3dsparserepresentationconstruction} is trained for 2 days on 32 A800 GPUs; the VecSet-DiT~\citep{zhang20233dshape2vecset,peebles2023scalable} is trained for 15 days on 128 GPUs; and the part-aware DiT is trained for 15 days on 256 GPUs with a total batch size of 256.

For training the part-aware DiT, we use the AdamW~\citep{kingma2014adam} optimizer with a learning rate of 1e-4 and no weight decay. An exponential moving average (EMA) with a rate of 0.9999 is applied. The unconditional guidance probability is set to 0.1, and the timestep sampling schedule follows a Logit-Normal distribution with a mean of 0 and standard deviation of 1. The model is conditioned on DINOv2~\citep{oquab2023dinov2} image embeddings. At inference time, we sample with a classifier-free guidance~\citep{ho2021classifier} scale of 3.5 over 25 steps.

\subsection{Qualitative Experiments}
\label{exp:qualitative}
\textbf{Comparison with Concurrent Methods.} As shown in Fig.~\ref{fig:compare}, \name~consistently outperforms prior methods in visual quality, demonstrating notable improvements in geometric detail and surface accuracy. These gains highlight the effectiveness of our framework and the benefits of our efficient and high resolution design.

\textbf{Ablation of Part Attention.} As shown in Fig.~\ref{fig:window_attn}, Part Attention achieves quality comparable to full global attention, while 3D window attention results in significantly degraded geometry. This is because 3D window attention relies on fixed partitions, which often misalign with object semantics and lead to fragmented attention computation. In contrast, our Part Attention respects semantic boundaries and preserves geometric continuity, enabling both accurate surface modeling and efficient computation. Note that both the Full Attention and 3D Window Attention models are finetuned from the Part Attention checkpoints for sufficient training steps.

\input{tabs/user}

\subsection{Quantitative Experiments}
\textbf{User Study.} We conducted user studies with 36 participants to evaluate the effectiveness of \name~and the proposed Part Attention. Each participant was presented with a set of image–3D mesh pairs and asked to choose the result that best matched the image in terms of overall quality and fidelity. Three sets of comparisons were conducted:

\begin{itemize}
    \item \textbf{(a) Comparison with Other Methods:} 20 image–mesh pairs comparing \name~with Direct3D-S2 and the commercial model A.
    \item \textbf{(b) Full Attention vs. Part Attention:} 10 image–mesh pairs comparing \name~with a variant where Part Attention is replaced by Full Attention.
    \item \textbf{(c) 3D Window vs. Part Attention:} 10 image–mesh pairs comparing \name~with a variant where Part Attention is replaced by 3D Window Attention.
\end{itemize}

\input{tabs/speed}

As shown in Tab.\ref{table:user}, \name~achieves the highest user preference in (a), with 68.5\% of selections—substantially outperforming existing state-of-the-art methods. In (b), Part Attention attains comparable preference to full attention, confirming its efficiency advantage without compromising quality. In (c), users clearly favored Part Attention over 3D window attention (63.7\% vs. 2.1\%), suggesting that our design better maintains geometric continuity and semantic alignment.

\textbf{Acceleration Benefits of Part Attention.}
We evaluate the acceleration benefits of Part Attention under the unified use of FlashAttention-2. As shown in Tab.~\ref{table:speedup}, both Part Self Attention and Cross Attention significantly reduce computation cost. Consequently, the overall training and inference pipeline is also accelerated. Depending on the number of active voxels, the full attention baseline often requires over 15 minutes to generate a single mesh—an impractical cost—whereas our pipeline averages just 4 minutes per sample.

%% file: tabs/user.tex
\begin{table*}[t]
  \centering
  \caption{\textbf{User Study.} Each table reports selection rates from user studies conducted on image–3D mesh pairs. Participants were asked to choose the result that best matches the image and exhibits the highest quality.}
  \vspace{2mm}

  \begin{minipage}[t]{\textwidth}
    \centering
    (a) Comparison with Other Methods\\
    \begin{tabular}{@{}lccc@{}}
      \toprule
      \textbf{Model} & Direct3D-S2 & Commercial Model A & Ours \\
      \midrule
      \textbf{Select.} & 7.2\% & 24.3\% & \textbf{68.5\%} \\
      \bottomrule
    \end{tabular}
  \end{minipage}
  \vspace{3mm}

  \begin{minipage}[t]{0.48\textwidth}
    \centering
    (b) Full Attention vs. Part Attention\\
    \begin{tabular}{@{}lccc@{}}
      \toprule
      \textbf{Model} & Ours-Full & Ours & No Pref. \\
      \midrule
      \textbf{Select.} & 12.4\% & 8.9\% & \textbf{78.7\%} \\
      \bottomrule
    \end{tabular}
  \end{minipage}
  \hfill
  \begin{minipage}[t]{0.48\textwidth}
    \centering
    (c) 3D Window vs. Part Attention\\
    \begin{tabular}{@{}lccc@{}}
      \toprule
      \textbf{Model} & Ours-Naive & Ours & No Pref. \\
      \midrule
      \textbf{Select.} & 2.1\% & \textbf{63.7\%} & 34.2\% \\
      \bottomrule
    \end{tabular}
  \end{minipage}

  \label{table:user}
\end{table*}

%% file: tabs/speed.tex
\begin{table}[h]
\centering
\caption{\textbf{Efficiency Comparison.} Part Attention significantly accelerate computation over full attention. Substituting most DiT layers with Part Attention yields a substantial speed-up in both training and inference. All attention blocks are implemented with FlashAttention-2~\citep{dao2023flashattention}.}
\vspace{2mm}
\begin{tabular}{@{}lcccc@{}}
\toprule
 & Part Self Attention & Part Cross Attention & DiT Training & DiT Inference \\
\midrule
\textbf{Speedup Rate}   & 6.7$\times$ & 4.1$\times$ & 3.1$\times$ & 3.3$\times$ \\
\bottomrule
\end{tabular}
\label{table:speedup}
\end{table}

%% file: 6_conclu.tex
\section{Conclusion}
In this work, we aim to address the computational bottlenecks of sparse voxel-based 3D generation. We identify the inefficiency of traditional two-stage pipelines
and propose a more scalable alternative by first generating a coarse mesh using VecSet and then voxelizing it into sparse voxels for refinement. To further alleviate the cost of full attention in the second stage, we introduce Part Attention—a geometry-aware attention mechanism that restricts computation to semantically coherent part groups. This design preserves detail while significantly reducing computation. To support this, we construct a large-scale, efficient part annotation pipeline, enabling high-quality part annotation at scale. Through extensive experiments, we demonstrate that our framework \name~achieves state-of-the-art visual quality with substantially improved efficiency in both training and inference, making high-resolution 3D generation more tractable and practical.